\pdfoutput=1
\documentclass{article}

\PassOptionsToPackage{numbers, compress}{natbib}

 \usepackage[main, final]{neurips_2026}

\usepackage[utf8]{inputenc} %
\usepackage[T1]{fontenc}    %
\usepackage{url}            %
\usepackage{booktabs}       %
\usepackage{amsfonts}       %
\usepackage{nicefrac}       %
\usepackage{microtype}      %
\usepackage{xcolor}         %
\usepackage[colorlinks=true, urlcolor=blue, linkcolor=blue, citecolor=blue]{hyperref}       %

\usepackage{todonotes}
\usepackage{siunitx}
\usepackage{amsmath}
\usepackage{enumitem}
\usepackage{cleveref}
\usepackage{subcaption}
\usepackage{multirow}
\usepackage{pifont}
\usepackage{colortbl}
\usepackage{threeparttable}
\usepackage{placeins}
\usepackage{booktabs}
\usepackage[table]{xcolor}
\usepackage{subcaption}
\usepackage{fontawesome}
\usepackage{float}
\usepackage{tikz}
\usetikzlibrary{plotmarks}
\definecolor{scannetblue}{HTML}{2166AC}
\definecolor{rscanred}{HTML}{D6604D}

\newcommand{\cmark}{\ding{51}}
\newcommand{\xmark}{\ding{55}}
\definecolor{chosengray}{RGB}{225,225,225}

\definecolor{bestperf}{HTML}{D1C4E9}   %
\definecolor{secondperf}{HTML}{FFE0B2} %

\newcommand{\best}{\cellcolor{bestperf}\bfseries}
\newcommand{\second}{\cellcolor{secondperf}}

\title{CrossWeave: Cross-Modal Scene and Instance Retrieval from Sparse 2D-3D Alignment}

\author{%
\textbf{Aadith Warrier}$^{1}$ \quad
\textbf{Gnana Prakash Punnavajhala}$^{1}$ \quad
\textbf{Siddharth Tourani}$^{2}$ \\
\textbf{Muhammad Haris Khan}$^{3}$ \quad
\textbf{Avinash Sharma}$^{4}$ \quad
\textbf{Madhava Krishna}$^{1}$\\[1.6ex]
$^{1}$IIIT Hyderabad \quad
$^{2}$University of Heidelberg \quad
$^{3}$MBZUAI \quad
$^{4}$IIT Bombay\\[1.6ex]
\faGlobe\;\href{https://crossweave3d.github.io/}{\texttt{crossweave3d.github.io}} \quad \faGithub\;\href{https://github.com/crossweave3d/crossweave}{\texttt{github.com/crossweave3d/crossweave}}
}

\begin{document}

\maketitle

\begin{abstract}
Cross-modal scene and instance retrieval aims to retrieve a corresponding 3D environment or object given query 2D images, serving as a foundational mechanism for spatial reasoning in robotics and AR/VR. Existing methods achieve this by learning alignment only at the level of whole scenes or objects, which provides no supervision for the fine-grained local structure within them. In this work, we show that supervision on a set of sparse 2D-3D correspondences, without scene- or instance-level contrastive objectives, is sufficient to train a single model for both scene- and instance-level retrieval. Specifically, we propose \textbf{CrossWeave}, a novel framework for cross-modal retrieval that learns sparse patch-point alignment, which alone outperforms existing methods. An optimal transport-based aggregator over the aligned features gives further gains by clustering structurally redundant regions, freeing descriptor capacity for semantically distinct local features. CrossWeave achieves state-of-the-art performance on both scene- and instance-level retrieval on ScanNet and 3RScan with a \textit{single model}, whereas existing methods require separate models for each task. Our results suggest that whole-scene encoding is unnecessary for cross-modal retrieval and that sparse local correspondence supervision is a more effective and flexible alternative.
\end{abstract}

\section{Introduction}

Within autonomous navigation and mapping, processing continuous streams of image and point cloud data is a major computational bottleneck. Scene and instance retrieval provide a first-pass spatial reasoning step, enabling location- and object-specific downstream tasks without exhaustive search over large databases or 3D maps. Such coarse reasoning is often sufficient for identifying object presence, recognizing the current environment, or initializing fine-grained localization and loop closure in SLAM. Cross-modal retrieval is especially valuable when the map is built using a richer sensor modality than the query device, such as a camera-equipped robot localizing against a LiDAR-built map, or an AR/VR device anchoring content to a pre-scanned environment.

Formally, \textbf{Cross-Modal Scene Retrieval (CMSR)} is defined as retrieving, from a database of aligned 3D point cloud sub-maps, the sub-map corresponding to query RGB image(s). The related task of \textbf{Cross-Modal Instance Retrieval (CMIR)} is defined analogously: given RGB image(s) of an object, the system retrieves the corresponding 3D point cloud instance from a database of segmented object point clouds. CMSR remains an emerging problem, with few established methods for bridging the modality gap between 2D images and 3D maps. Existing approaches such as \textit{CrossOver}~\cite{Sarkar_2025_CVPR} and \textit{SceneGraphLoc}~\cite{miao2024scenegraphloc} rely on scene- or object-level supervision: CrossOver aligns modalities with separate scene- and instance-level contrastive objectives and requires a separate model for each task, while SceneGraphLoc matches image patches to the object nodes of a multimodal scene graph. In both cases, descriptors are learned using contrastive objectives similar to those used in visual place recognition~\cite{arandjelovic2016netvlad, izquierdo2024optimal}. However, both methods supervise alignment at the level of whole scenes or objects rather than individual 3D points.

To address this, we propose \textbf{CrossWeave}, a cross-modal retrieval framework that learns from sparse local scene regions during training instead of aligning full scenes. CrossWeave introduces an image-patch-to-3D-point contrastive objective that preserves local cross-modal correspondences. It further aligns corresponding image-point cloud pairs using modality-specific descriptors obtained by aggregating local features, without requiring full-scene alignment. Using this architecture, we show that a single model trained with these losses, without any scene- or instance-level supervision, supports both scene- and object-level retrieval on the 3RScan and ScanNet benchmarks. We further demonstrate that patch-point alignment is the primary driver of instance retrieval performance, and improves scene retrieval when combined with learned aggregation.

\paragraph{Contributions: } 
\begin{itemize}[leftmargin=2em]
    \item We propose \textbf{CrossWeave}, a novel multi-modal framework that performs both CMSR and CMIR with a single model, using a dual-contrastive loss at both the local patch-point level and the image-point cloud level, but not a scene/instance-level contrastive loss. This allows improved retrieval when compared to methods that encode the whole 3D map either as global point cloud features or scene graphs.
    \item We show that adding local patch-point alignment consistently improves both CMSR and CMIR over training on aggregated descriptors alone, with the largest gains on fine-grained instance retrieval.
    \item We achieve state-of-the-art performance on 3RScan and ScanNet, outperforming competing baselines by 21 points on scene retrieval and 8 points on instance retrieval on ScanNet.
\end{itemize}

\section{Related Work}

\subsection{Multi-modal Representation Learning} 
Multi-modal representation learning aims to align heterogeneous data sources within a shared embedding space. In this setting, contrastive learning has become the dominant paradigm. CLIP~\cite{pmlr-v139-radford21a} demonstrated that large-scale contrastive training on paired image-text data yields rich representations capable of cross-modal understanding and strong zero-shot generalization to various downstream tasks. Building on this foundation, a series of works extended cross-modal alignment to the image-point cloud~\cite{Zhang_2022_CVPR, Zhu_2023_ICCV} and text-image-point cloud modalities~\cite{Xue_2023_CVPR, Xue_2024_CVPR, NEURIPS2023_8c7304e7} with Uni3D~\cite{zhou2023uni3d} scaling model and dataset sizes used for training. Subsequently, ImageBind~\cite{Girdhar_2023_CVPR} performed alignment between six modalities by leveraging images as a hub and demonstrated emergent relationships between untrained pairs. Similarly, PointBind~\cite{guo2023pointbindpointllmaligning} incorporated the 3D modality and performed various downstream tasks by augmenting with an LLM. However, these methods remained restricted to object-level point clouds. CLIP$^2$~\cite{Zeng_2023_CVPR} extended multi-modal representation learning to scene-level point clouds, yet remained focused on semantic understanding rather than retrieval. Overall, prior work focuses on cross-modal alignment for semantic understanding, with comparatively less emphasis on retrieval-oriented representation design for 3D scenes.

\subsection{Uni-modal and Cross-modal Retrieval}
Retrieval methods aim to identify relevant database entries given a query, typically represented by a global descriptor. While early methods aggregated local features via hand-crafted clustering~\cite{vlad}, later works introduced learned aggregation like NetVLAD~\cite{arandjelovic2016netvlad} to preserve distributional structure. Uni-modal visual~\cite{arandjelovic2016netvlad, keetha2023anyloc} and LiDAR-based~\cite{Uy_2018_CVPR, logg3d, btc_lpr, puligilla2024lip, zheng2023i2p} place recognition are widely studied along these lines. In the cross-modal domain, image-text retrieval is well-established~\cite{cao2022image}, heavily driven by contrastive joint-embedding alignment~\cite{pmlr-v139-radford21a, lu2023cross, schall2024optimizing}. Extending beyond 2D, recent works bridge textual queries with 3D point clouds via part-to-word matching~\cite{tang2023parts2words} and multi-modal consistencies~\cite{ren2025enhanced, feng2025pointcloud}. Concurrently, scene graphs are used to capture explicit structural relations for cross-modal text-to-image retrieval, evolving from basic query parsing~\cite{johnson2015image} to advanced message passing and attribute alignment~\cite{duan2021hsgmp, wang2025sgg}. Despite these advances, cross-modal scene and instance retrieval remains largely underexplored, particularly for indoor environments mapped between images and 3D point clouds.

The most closely related work to ours is CrossOver~\cite{Sarkar_2025_CVPR}, which trains for cross-modal alignment across text, images, floorplans, CAD models, and point clouds at both scene and instance levels, enabling retrieval across any pair of these modalities at multiple levels of granularity. We highlight two design aspects of prior work most relevant to our setting. First, instance- and scene-level retrieval rely on separate task-specific checkpoints, requiring distinct models for different granularities and limiting flexibility at inference time. Second, scene-level representations are constructed by averaging per-frame features (CLS and mean patch tokens) across frames for images and mean pooling point features for point clouds, which implicitly assigns equal weight to all local features, potentially losing fine-grained structural information critical for distinguishing visually similar scenes. In contrast, our method uses a unified model that learns fine-grained patch-to-point correspondences, which can be flexibly aggregated at inference time to support both instance- and scene-level retrieval. Furthermore, our aggregation assigns local features to shared cluster centers, capturing their distributional structure rather than averaging them away.

\section{Methodology}
\begin{figure}
    \centering
    \includegraphics[width=\textwidth]{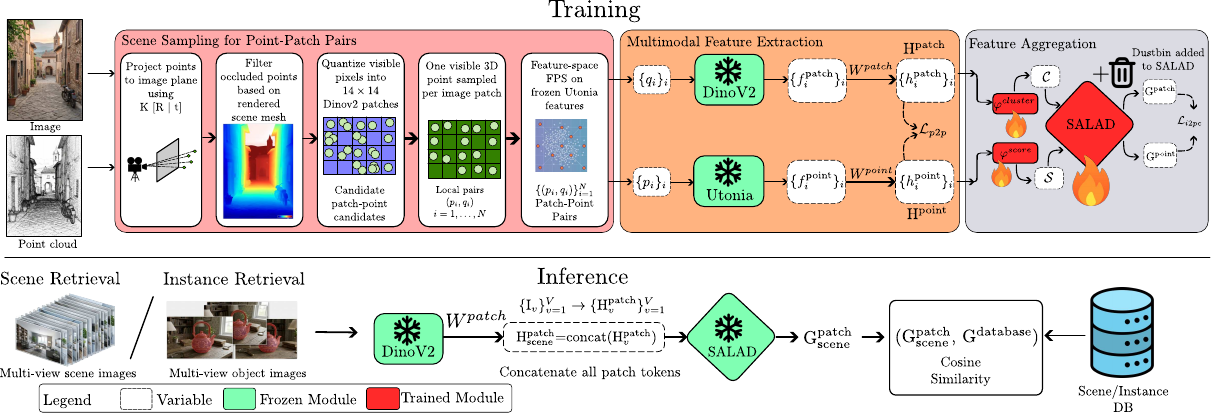}
    \caption{\textbf{Overview of CrossWeave.} CrossWeave learns cross-modal retrieval representations from sparse image patch-3D point correspondences rather than scene- or instance-level labels. \textbf{Scene Sampling:} Given calibrated RGB-D observations and a reconstructed 3D map, visible patch-point pairs are obtained by projecting 3D points into the image plane, filtering occluded points using rendered depth, quantizing valid pixels into DINOv2 patch tokens, and selecting a compact semantically diverse subset using feature-space farthest point sampling. \textbf{Feature Extraction:} Frozen encoders extract local patch and point features, which are projected by learnable projection matrices $W^{\mathrm{patch}}$ and $W^{\mathrm{point}}$ into a shared embedding space. \textbf{Feature Aggregation:} A shared SALAD aggregator then maps variable-sized sets of local features into fixed-dimensional descriptors. Training combines a local patch-point $\mathcal{L}_{p2p}$ contrastive loss with an aggregated image-point-cloud contrastive loss $\mathcal{L}_{i2pc}$, preserving fine-grained alignment while organizing local evidence into retrieval-ready descriptors. At inference, the same encoders and projection heads serve both tasks: SALAD descriptors are used for scene retrieval and mean-pooled local features for instance retrieval.}
    \label{fig:overview}
\end{figure}

The design philosophy behind CrossWeave is to both: effectively utilize the redundancy in the 3D scene structure and not lose fine-grained details due to the aggregation process over the local features to obtain modality-specific descriptors. \Cref{fig:overview} provides an overview of CrossWeave. 

\subsection{Architecture and Training}  \label{subsec:training}
In keeping with the convention in previous CMSR literature, we assume the 3D structure consists of individual RGB-D scans aligned into a single 3D map and not the individual scans themselves. 

\paragraph{Scene Sampling for Patch-Point Pair Generation} To exploit the redundancy present in both image and point-cloud observations, we avoid encoding all 3D points visible in an image and instead construct a compact set of informative patch-point correspondences. Given a point cloud \(P\), we first project each 3D point into the image plane using the camera intrinsics \(K\) and camera pose \([R \mid t]\). Since projection alone does not guarantee that a point is actually visible from the camera viewpoint, we apply an explicit visibility constraint using the depth map rendered from the reconstructed mesh. A projected point is retained only if its depth lies within a small tolerance \(\epsilon\) of the observed depth at the corresponding pixel. This removes occluded points and ensures that the resulting correspondences reflect physically visible image-geometry associations. These correspondences are the only supervision used during training, and no scene or instance labels are required.

The remaining valid pixel-point pairs are then quantized into image patch indices using a patch size of \(14\), following the patchification scheme of DINOv2~\cite{dinov2}. For each image patch, we randomly sample a single visible 3D point, producing a one-to-one patch-point mapping. This step reduces the strong local redundancy caused by multiple 3D points projecting to the same visual token, while preserving the spatial correspondence required for patch-level cross-modal supervision.

Finally, we further downsample these patch-point pairs using Farthest Point Sampling (FPS)~\cite{fpsEldar} in feature space rather than Euclidean space. Specifically, FPS is applied to the point features extracted by our frozen 3D encoder, Utonia~\cite{zhang2026utonia}, a recent point-cloud feature extractor based on the Point Transformer v3 architecture~\cite{wu2024point}. Unlike geometric FPS, which primarily encourages spatial coverage, feature-space FPS encourages semantic diversity by selecting points that are maximally separated in the learned embedding space. This prevents the sampled set from being dominated by large or frequently observed objects in the field of view, yielding a more balanced set of local correspondences for contrastive learning. The output of this scene sampling is a set of point-patch pairs $\{p_i \in \mathbb{R}^3, q_i \in \mathbb{R}^{14 \times 14} \}, i=1 \rightarrow N$.

\paragraph{Feature Extraction}
The sampled point-patch pairs are passed on to pre-trained modality-specific encoders. For the image patch $q_i$, we use the DINOv2-L variant for generating per-patch features $f_i^{\text{patch}}$. For the 3D point $p_i$, we use Utonia to extract per-point features $f_i^{\text{point}}$ from the point cloud. Subsequently, both features are $\ell_2$ normalized and projected into a shared embedding space using learnable projection matrices $W_{\text{patch}}$ and $W_{\text{point}}$.
This results in per modality aligned features $H^{patch} = \{h_i^{\text{patch}}\}_{i=1}^N$ and $H^{point} = \{h_i^{\text{point}}\}_{i=1}^N$ per image. 

\paragraph{Feature Aggregation via Optimal Transport}
For each image-point-cloud sample, we aggregate the \(N\) sampled patch-point correspondences into fixed-dimensional descriptors using a SALAD~\cite{izquierdo2024optimal} aggregation layer. Here, \(N\) denotes the number of retained patch-point pairs after visibility filtering and sampling, while \(K\) denotes the number of latent aggregation clusters used by SALAD. The goal of this aggregation is to summarize a compact set of local cross-modal correspondences into a single descriptor, while capturing discriminative and recurring patterns in the cross-modal data. Importantly, aggregation is performed only over the \(N\) sampled correspondences from one image-point-cloud pair, rather than over all points in a full scene or features of an object instance.

To complete the mapping from patch-point level local features ($H^{point}$, $H^{patch}$) to image-point cloud level features,  SALAD uses two MLPs shared across both modalities. The first MLP, \(\varphi^{\text{cluster}}\), maps each local feature to a cluster feature, while the second MLP, \(\varphi^{\text{score}}\), predicts its assignment scores to the \(K\) latent clusters:
\[
C = \varphi^{\text{cluster}}(H) \in \mathbb{R}^{N \times d_{\text{cl}}}, \qquad
S = \varphi^{\text{score}}(H) \in \mathbb{R}^{N \times K}.
\]
Here, \(d_{\text{cl}}\) is the dimensionality of the transformed local feature used inside the aggregation layer. Thus, \(C_i\), the \(i\)-th row of $C$, is the feature content contributed by the \(i\)-th local correspondence, while \(S_{i,k}\) measures how strongly it should be assigned to the \(k\)-th latent cluster.

Following SALAD~\cite{izquierdo2024optimal}, the score matrix \(S\) is augmented with a learnable dustbin parameter \(\phi\) and passed through the Sinkhorn algorithm~\cite{cuturi2013sinkhorn} for three iterations. This produces a balanced soft-assignment matrix $\mathcal{P} \in \mathbb{R}^{K \times N}$, after removing the dustbin assignments. The dustbin allows low-confidence or uninformative local features to be absorbed rather than being forced into one of the \(K\) clusters, which improves robustness when some sampled correspondences are noisy or weakly discriminative. Each cluster descriptor \(V_k\) is then computed as a weighted sum of the transformed local features assigned to the \(k\)-th cluster.
Thus, each cluster summarizes the subset of local evidence associated with the \(k\)-th latent concept. The vectors \(V_k\) are individually \(\ell_2\)-normalized, concatenated, and then globally \(\ell_2\)-normalized to obtain the final descriptor.
\[
G = \operatorname{norm}\Big(
\operatorname{concat}\big(
\operatorname{norm}(V_1), \dots, \operatorname{norm}(V_K)
\big)
\Big)
\in \mathbb{R}^{K d_{\text{cl}}}.
\]
Applying this aggregation independently to \(H^{\text{patch}}\) and \(H^{\text{point}}\) yields the modality-specific descriptors \(G^{\text{patch}}\) and \(G^{\text{point}}\), which are later used for cross-modal contrastive supervision.

\paragraph{Dual-Objective Contrastive Learning}
CrossWeave is trained with two complementary contrastive objectives, both based on InfoNCE~\cite{oord2018representation, Musgrave2020PyTorchML}. The first objective, the local patch-point loss \(\mathcal{L}_{\text{p2p}}\), is computed over all \(M = B \times N\) patch-point correspondences in a batch, where \(B\) is the batch size and \(N\) is the number of sampled correspondences per image. This loss directly aligns local visual and geometric features, encouraging each image patch to be close to its corresponding 3D point while treating other patch-point pairs in the batch as negatives. Since it is defined only at the correspondence level, it does not require scene-level or instance-level annotations.

The second objective, the image-point-cloud loss \(\mathcal{L}_{\text{i2pc}}\) realized as the InfoNCE loss as well, is applied to the SALAD-aggregated descriptors of each image-point-cloud pair in the batch. This loss encourages the aggregated image descriptor and the corresponding aggregated point-cloud descriptor to occupy a shared retrieval space. Unlike full-scene supervision, this alignment is derived from descriptors built only from the sampled local correspondences, preserving the focus on sparse but discriminative cross-modal evidence.

The final training objective combines the local and aggregated contrastive losses, \(\mathcal{L} = \mathcal{L}_{\text{p2p}} + \mathcal{L}_{\text{i2pc}}\). Together, these objectives allow CrossWeave to learn retrieval-oriented representations from sparse patch-point correspondences alone. The local loss preserves fine-grained cross-modal alignment, while the aggregated loss organizes these local cues into image-level and point-cloud-level descriptors. 

\subsection{Inference}

Although CrossWeave is trained only on sparse local patch-point correspondences, it can be applied directly to both scene-level and instance-level retrieval at inference. This follows from the input-agnostic design of the model. The projection layers are applied independently to each local image or point feature, and the SALAD aggregator~\cite{izquierdo2024optimal} maps any variable-sized set of local features to a fixed-dimensional descriptor. Thus, the same network can aggregate a small set of training correspondences, a complete scene point cloud, or a segmented object instance without architectural modification. SALAD produces a descriptor \(G \in \mathbb{R}^{K d_{\text{cl}}}\), where \(K\) is the number of learned clusters and \(d_{\text{cl}}\) is the dimensionality of each cluster vector.

\paragraph{Scene Retrieval}
For scene-level retrieval, the full scene point cloud is encoded using Utonia, projected into the shared embedding space, and aggregated with SALAD to obtain a point-cloud descriptor \(G^{\text{point}}\). Let the scene be observed by \(V\) images, where the \(v\)-th image produces \(N_v\) DINOv2 patch tokens. We concatenate the patch tokens from all views before aggregation:
\[
H^{\text{patch}}
=
\operatorname{concat}
\left(
H^{\text{patch}}_1,\ldots,H^{\text{patch}}_V
\right)
\in
\mathbb{R}^{(\sum_{v=1}^{V} N_v) \times d}.
\]
These concatenated multi-view patch features are passed through the image projection layer and SALAD aggregator to produce the image-side scene descriptor \(G^{\text{patch}}\). Retrieval is then performed by ranking database descriptors according to their cosine similarity with the query.

\paragraph{Instance Retrieval}
Instance-level retrieval restricts the inputs to a segmented object instance. The scene point cloud is replaced by the instance point cloud, and each query view is an RGB image together with the instance mask, of which only the patch tokens covered by the mask are used. For instance retrieval we use the local features directly instead of the SALAD descriptor: the projected patch features from all query views and the projected instance point features are each mean-pooled, and the two resulting descriptors are compared using cosine similarity. We use local features because they give better instance retrieval than the SALAD descriptor on both datasets (\Cref{tab:ablation}).

\subsection{Implementation Details} \label{subsec:implementation_details}
CrossWeave is implemented in PyTorch~\cite{paszke2019pytorch} and trained on a single NVIDIA A6000 GPU using mixed-precision \(\mathrm{bfloat16}\) training. Training requires 17.1 GB of VRAM and about 21 GPU hours. We use the AdamW optimizer~\cite{loshchilov2017decoupled} with a learning rate of \(10^{-3}\), weight decay of \(10^{-2}\), and train for 30 epochs with a batch size of 16. During training, each sample contains 32 patch-point correspondences. DINOv2-L produces 1024-dimensional image patch features, and Utonia produces 1224-dimensional point features. These are projected into a shared 1024-dimensional embedding space using modality-specific projection matrices \(W_{\text{patch}}\) and \(W_{\text{point}}\). The SALAD aggregator uses \(K=24\) cluster vectors with \(d_{\text{cl}}=128\), producing a final descriptor of dimension $K \times d_{\text{cl}} = 24 \times 128 = 3072$. We measure inference time on a single NVIDIA A6000 GPU using 10 query images of resolution $3 \times 518 \times 518$ and a point cloud of $50{,}000$ points. The RGB forward pass (DINOv2 + SALAD) runs in $118.69 \pm 1.87$ ms and the point cloud forward pass (Utonia + SALAD) runs in $332.15 \pm 4.52$ ms, with a combined peak memory usage of $5.24$ GB. In practice, the point cloud database descriptors are computed offline and cached, so only the RGB forward pass is required at query time.

\section{Experiments}

\subsection{Datasets}
\label{subsec:datasets}

\paragraph{ScanNet~\cite{dai2017scannet}}
ScanNet is a large-scale indoor dataset containing posed RGB-D frames from 707 scenes and 1513 scans, with multiple runs per scene. The dataset provides dense 3D meshes, with semantic and instance annotations for RGB frames and reconstructed meshes. We use the standard splits of 565 scenes / 1201 scans for training and 142 scenes / 312 scans for validation.

\paragraph{3RScan~\cite{Wald_2019_ICCV}}
3RScan contains 1482 scans of 478 indoor scenes, with posed RGB-D frames, dense 3D meshes, and semantic and instance annotations. We use 384 scenes / 1177 scans for training and 46 scenes / 157 scans from the validation split for evaluation following~\cite{lu2023cross}. 

Additional dataset details are provided in \Cref{sec:more_implementation_details}.

\subsection{Tasks and Metrics}

\paragraph{Metrics}
For both scene-level and instance-level retrieval, we report \emph{Top-$k$ accuracy}, defined as the percentage of queries for which the correct scene or instance appears among the top-$k$ retrieved results. For instance retrieval, we additionally report \emph{Recall@$k$\%} (R@$k$\%), which measures the percentage of scenes for which at least $k$\% of the query instances are correctly retrieved.

\paragraph{Scene Retrieval}
In the scene retrieval task, the query consists of 10 RGB views of a scene, selected following the protocol of CrossOver~\cite{Sarkar_2025_CVPR}, and the database consists of 3D point clouds reconstructed from scanned scenes. A retrieval is considered correct if the model retrieves the point cloud corresponding to the queried scene. The database includes both original scans and re-scans, where the same environment may be captured again after objects have been displaced.

\paragraph{Instance Retrieval}
In the instance retrieval task, the query consists of multiple RGB views of an object instance, each with its instance mask. The model must retrieve the corresponding 3D object point cloud from the instance point clouds of the scene the query was taken from. The task therefore evaluates fine-grained object matching within a known environment. Top-$k$ accuracy is computed within each scene and averaged over all validation scenes.

\subsection{Baselines}

Our primary baseline is CrossOver~\cite{Sarkar_2025_CVPR}. We evaluate two publicly released CrossOver checkpoints for each task: one trained on the same datasets as our method, and another checkpoint of the same model trained on additional data, which we denote CrossOver$^\dagger$. To ensure that performance differences are not caused by the choice of 3D encoder, we also retrain CrossOver on its original preprocessed training data after replacing its 3D backbone with Utonia~\cite{zhang2026utonia}; we refer to this variant as \textit{CrossOver (Updated)}. All methods are evaluated with CrossOver's evaluation protocol for a fair comparison.

\section{Results} \label{sec:results}

\subsection{Cross-Modal Scene Retrieval}

\begin{table*}[t]
\centering
\footnotesize{
\caption{Comparison of scene retrieval performance on ScanNet and 3RScan. 
$^\dagger$ Trained on additional datasets (ArkitScenes~\cite{baruch2021arkitscenes} + MultiScan~\cite{mao2022multiscan}). 
\colorbox{blue!20}{\textbf{Best}} and
\colorbox{orange!20}{Second-best} results.}
\label{tab:scene_results}
\begin{tabular}{l ccc ccc}
\toprule
& \multicolumn{3}{c}{\textbf{ScanNet}} 
& \multicolumn{3}{c}{\textbf{3RScan}} \\
\cmidrule(lr){2-4} \cmidrule(lr){5-7}
& \multicolumn{3}{c}{Scene Recall} 
& \multicolumn{3}{c}{Scene Recall} \\
\cmidrule(lr){2-4} \cmidrule(lr){5-7}
\textbf{Method}
& Top-1 & Top-5 & Top-10
& Top-1 & Top-5 & Top-10 \\
\midrule

Scene CrossOver
  & 22.44 & 61.86 & 75.32
  & 19.11 & 58.60 & 77.07 \\

Scene CrossOver$^\dagger$ 
  & 19.87 & 58.01 & 74.36
  & 17.83 & 51.59 & 73.25 \\

Scene CrossOver (Updated)
  & \second{40.38} & \second{90.06} & \second{96.47}
  & \second{23.57} & \second{66.88} & \second{84.71} \\

\textbf{Ours} 
  & \best{61.54} & \best{99.36} & \best{100.00}
  & \best{46.50} & \best{85.99} & \best{92.36} \\

\bottomrule
\end{tabular}%
}
\end{table*}

Scene retrieval results are reported in~\Cref{tab:scene_results}. CrossWeave consistently outperforms all baselines on both ScanNet and 3RScan across all retrieval ranks. On ScanNet, CrossWeave achieves a Top-1 recall of $61.54$, improving over CrossOver (Updated) by $21.16$ points. The gains remain strong at higher ranks, where CrossWeave reaches $99.36$ Top-5 recall and perfect Top-10 recall. These results indicate that the correct scene is almost always ranked among the first few retrieved candidates.

The improvement is particularly meaningful because CrossOver (Updated) uses the same Utonia backbone. This suggests that the performance gain does not simply come from stronger 3D encoding, but from our training strategy based on sparse local patch-point supervision. By aligning local image patches with corresponding 3D points, CrossWeave learns descriptors that preserve fine-grained geometric and visual correspondences. This helps disambiguate scenes with similar global layouts, such as corridors, bedrooms, or repeated furniture arrangements, where purely global descriptors can collapse visually similar environments.

On 3RScan, CrossWeave shows the same trend. It achieves a Top-1 recall of $46.50$, compared to $23.57$ for CrossOver (Updated), corresponding to a $22.93$ point improvement. CrossWeave also improves Top-5 recall from $66.88$ to $85.99$ and Top-10 recall from $84.71$ to $92.36$. These results demonstrate that local patch-point alignment generalizes across datasets and remains effective under the more challenging scene variations present in 3RScan. We also observe that CrossOver$^\dagger$ does not improve over the base CrossOver model, suggesting that additional training data alone is insufficient without a training objective that explicitly encourages cross-modal local alignment.

\begin{figure}[t]
    \centering
    \includegraphics[width=\linewidth]{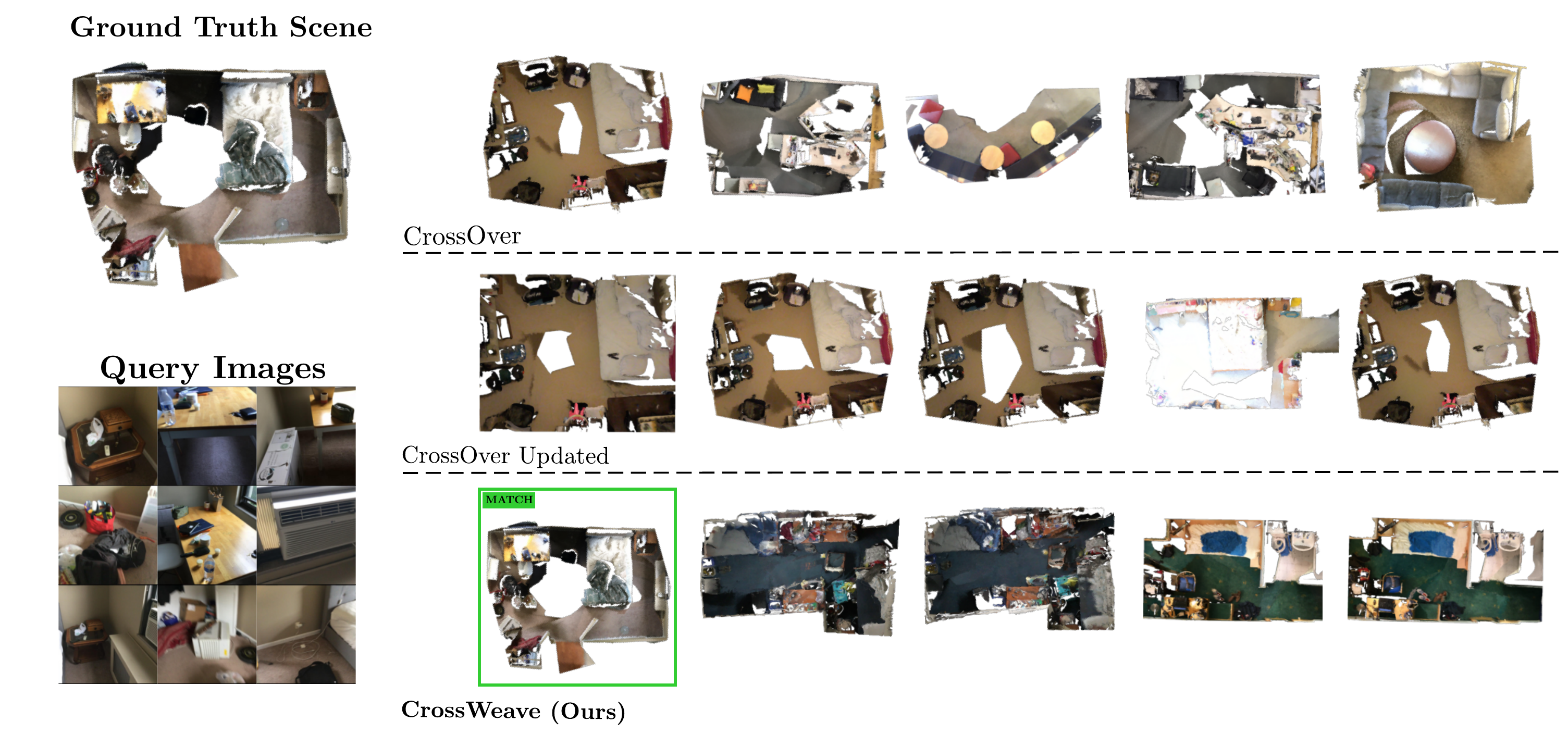}
    \caption{
    Qualitative comparison of cross-modal scene retrieval on ScanNet. Given a set of query RGB images, each model retrieves the most similar 3D scenes from the database. The ground-truth scene is shown on the left, and retrieved scenes are ordered by similarity from left to right. CrossOver and CrossOver (Updated) fail to retrieve the correct scene within the top five results, while CrossWeave retrieves it at rank 1. The remaining retrieved scenes are also semantically consistent, indicating that CrossWeave learns descriptors that capture both local cross-modal correspondences and global scene context.
    }
    \label{fig:scene_qualitative}
\end{figure}

\Cref{fig:scene_qualitative} shows a representative qualitative example on ScanNet. Given a set of query images, CrossWeave retrieves the correct 3D scene at rank 1, whereas both CrossOver baselines fail to retrieve the ground-truth scene within the top five results. This behavior is consistent across different scene categories: even when the query images cover only partial or localized views, CrossWeave produces descriptors that remain discriminative enough to separate visually similar scenes. Additional qualitative examples are provided in~\Cref{sec:addl_viz}, and failure cases are analyzed in~\Cref{sec:failure}.

\subsection{Cross-Modal Instance Retrieval}

\begin{table*}[t]
\centering
\caption{Comparison of within-scene instance retrieval performance on ScanNet and 3RScan.
$^\dagger$ Trained on additional datasets (ArkitScenes~\cite{baruch2021arkitscenes} + MultiScan~\cite{mao2022multiscan}). 
\colorbox{blue!20}{\textbf{Best}} and
\colorbox{orange!20}{Second-best} results.}
\label{tab:scannet_3rscan_instance_results}
\footnotesize

\begin{minipage}[t]{0.49\textwidth}
\centering
\caption*{\textbf{ScanNet}}
\resizebox{\linewidth}{!}{%
\begin{tabular}{l cc ccc}
\toprule
& \multicolumn{2}{c}{Instance Retrieval}
& \multicolumn{3}{c}{Scene-Level Recall} \\
\cmidrule(lr){2-3} \cmidrule(lr){4-6}
\textbf{Method}
& Top-1 & Top-3
& R@25\% & R@50\% & R@75\% \\
\midrule

Instance CrossOver
  & 63.58 & 88.71
  & 96.79 & 80.77 & 31.09 \\

Instance CrossOver$^\dagger$
  & 63.42 & 88.55
  & 96.79 & 80.77 & 31.09 \\

Instance CrossOver (Updated)
  & \second{70.94} & \second{92.39}
  & \best{99.36} & \second{88.14} & \second{49.68} \\

\textbf{Ours}
  & \best{78.77} & \best{94.05}
  & \second{98.40} & \best{88.78} & \best{67.95} \\

\bottomrule
\end{tabular}%
}
\end{minipage}
\hfill
\begin{minipage}[t]{0.49\textwidth}
\centering
\caption*{\textbf{3RScan}}
\resizebox{\linewidth}{!}{%
\begin{tabular}{l cc ccc}
\toprule
& \multicolumn{2}{c}{Instance Retrieval}
& \multicolumn{3}{c}{Scene-Level Recall} \\
\cmidrule(lr){2-3} \cmidrule(lr){4-6}
\textbf{Method}
& Top-1 & Top-3
& R@25\% & R@50\% & R@75\% \\
\midrule

Instance CrossOver
  & 58.88 & 84.24
  & 90.45 & 71.97 & 28.03 \\

Instance CrossOver$^\dagger$
  & 57.09 & 84.58
  & 89.81 & 63.06 & 24.84 \\

Instance CrossOver (Updated)
  & \second{66.80} & \second{90.85}
  & \second{97.45} & \second{87.90} & \second{35.03} \\

\textbf{Ours}
  & \best{73.50} & \best{92.73}
  & \best{98.09} & \best{92.99} & \best{49.68} \\

\bottomrule
\end{tabular}%
}
\end{minipage}

\end{table*}

Instance retrieval results are reported in~\Cref{tab:scannet_3rscan_instance_results}. Each query instance is retrieved from the instances of its own scene.

On ScanNet, CrossWeave achieves the best instance retrieval performance, with a Top-1 recall of $78.77$ compared to $70.94$ for CrossOver (Updated), an improvement of $7.83$ points. CrossWeave also obtains the best Top-3 recall, improving from $92.39$ to $94.05$. Local patch-point supervision therefore improves fine-grained instance discrimination as well as scene matching.

Across scenes on ScanNet, the difference between methods becomes more pronounced at stricter recall thresholds. At R@25\%, CrossWeave and CrossOver (Updated) are both near saturation, achieving $98.40$ and $99.36$, respectively. At R@50\%, CrossWeave obtains the best result with $88.78$, slightly outperforming CrossOver (Updated). The largest gain appears at R@75\%, where CrossWeave reaches $67.95$ compared to $49.68$ for CrossOver (Updated), an improvement of $18.27$ points. This suggests that CrossWeave is especially effective when the task requires retrieving a larger fraction of all correct instances, rather than only identifying the easiest matches.

The same pattern is observed on 3RScan. CrossWeave achieves the best instance retrieval performance, with a Top-1 recall of $73.50$ and a Top-3 recall of $92.73$, outperforming CrossOver (Updated) by $6.70$ and $1.88$ points, respectively. CrossWeave also obtains the best scene-level recall at all thresholds, reaching $98.09$ at R@25\%, $92.99$ at R@50\%, and $49.68$ at R@75\%. The improvement at R@75\% is especially notable, where CrossWeave improves over CrossOver (Updated) by $14.65$ points.

\begin{figure}[t]
    \centering
    \includegraphics[width=\linewidth]{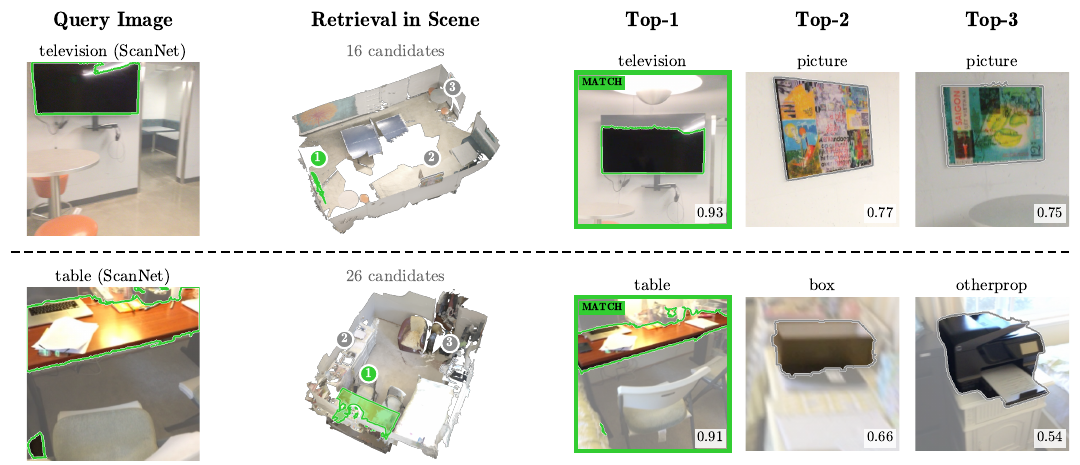}
    \caption{Qualitative instance retrieval on ScanNet. For each query view with its instance mask (left), CrossWeave ranks the instances of the same scene (middle) by cosine similarity. The Top-3 retrievals are shown with their similarity scores, and the correct match is outlined in green.}
    \label{fig:instance_qualitative}
\end{figure}

Overall, the instance retrieval results support the same conclusion as the scene retrieval experiments: sparse local patch-point alignment provides a stronger cross-modal training signal than instance-level supervision alone. While the baselines are often sufficient to retrieve a small number of easy matches, CrossWeave is more reliable under stricter retrieval criteria, where the model must recover a broader and more complete set of correct instances. \Cref{fig:instance_qualitative} shows two examples on ScanNet, where CrossWeave ranks the correct instance first among $16$ and $26$ candidates in the scene. Results with noisy 2D and 3D instance masks are reported in \Cref{subsec:degraded_masks}.

\subsection{Single View Scene Retrieval}

\Cref{tab:scenegraphloc} compares CrossWeave with SceneGraphLoc~\cite{miao2024scenegraphloc} and CrossOver~\cite{Sarkar_2025_CVPR} on 3RScan, where each query image is matched against $N$ candidate scenes, following the evaluation protocol of SceneGraphLoc. SceneGraphLoc uses ground-truth scene graphs at inference, whereas CrossWeave uses only a single query image and the raw point cloud of each candidate scene. With 10 candidates, CrossWeave matches SceneGraphLoc at Top-1 ($53.60$) and improves Top-5 recall from $81.90$ to $93.10$ and Top-10 recall from $92.80$ to $100.00$. Recall drops for all methods when the pool grows to 50 candidates. In this setting, CrossWeave is $2.70$ points behind SceneGraphLoc at Top-1 ($27.50$ vs. $30.20$), but leads by $10.00$ points at Top-5 and $15.30$ points at Top-10. CrossWeave also outperforms CrossOver in every setting, by between $7.60$ and $24.88$ points. Since CrossWeave needs no scene graph annotations at inference, it can be deployed where such annotations are not available.

\begin{table}[t]
\centering
\caption{Scene retrieval on 3RScan against SceneGraphLoc and CrossOver. Each query image is ranked against $N$ candidate scenes; Top-$k$: percentage of queries with the correct scene in the top $k$. $^\ddagger$ Uses ground-truth scene graphs at inference.
\colorbox{bestperf}{\textbf{Best}} and \colorbox{secondperf}{second-best} results are highlighted.}
\label{tab:scenegraphloc}
\footnotesize
\setlength{\tabcolsep}{6pt}
\renewcommand{\arraystretch}{1.15}
\begin{tabular}{l ccc ccc}
\toprule
& \multicolumn{3}{c}{\textbf{10 candidates}} & \multicolumn{3}{c}{\textbf{50 candidates}} \\
\cmidrule(lr){2-4} \cmidrule(lr){5-7}
\textbf{Method} & Top-1 & Top-5 & Top-10 & Top-1 & Top-5 & Top-10 \\
\midrule
SceneGraphLoc$^\ddagger$~\cite{miao2024scenegraphloc}
  & \best{53.60} & \second{81.90} & \second{92.80}
  & \best{30.20} & \second{50.20} & \second{61.20} \\
CrossOver~\cite{Sarkar_2025_CVPR}
  & 46.00 & 77.97 & 90.58
  & 18.69 & 39.16 & 51.62 \\
\textbf{Ours}
  & \best{53.60} & \best{93.10} & \best{100.00}
  & \second{27.50} & \best{60.20} & \best{76.50} \\
\bottomrule
\end{tabular}
\end{table}

\subsection{Ablation Studies}

\begin{table*}[t]
\centering
\caption{Ablation of training losses, aggregation and retrieval features.
\emph{Agg.}: aggregation used during training.
\emph{Feats}: \textsc{salad} (aggregated descriptors), \textsc{local} (mean-pooled patch and point features), or \textsc{hybrid} (\textsc{salad} for scene, \textsc{local} for instance retrieval).
\colorbox{bestperf}{\textbf{Best}} and \colorbox{secondperf}{second-best} results are highlighted; the main model is shaded and marked with $^\star$.}
\label{tab:ablation}
\footnotesize
\setlength{\tabcolsep}{6pt}
\renewcommand{\arraystretch}{1.15}
\begin{tabular}{l cc l ccc ccc}
\toprule
& & & & \multicolumn{3}{c}{\textbf{ScanNet}} & \multicolumn{3}{c}{\textbf{3RScan}} \\
\cmidrule(lr){5-7} \cmidrule(lr){8-10}
& & & & Scene & \multicolumn{2}{c}{Instance} & Scene & \multicolumn{2}{c}{Instance} \\
\cmidrule(lr){5-5} \cmidrule(lr){6-7} \cmidrule(lr){8-8} \cmidrule(lr){9-10}
\textbf{Agg.} & $\mathcal{L}_{i2pc}$ & $\mathcal{L}_{p2p}$ & \textbf{Feats}
& Top-1 & Top-1 & R@75\% & Top-1 & Top-1 & R@75\% \\
\midrule
Mean  & \xmark & \cmark & \textsc{local}  & 47.12 & \second{78.96} & \second{70.51} & 27.39 & 73.30 & 47.13 \\
Mean  & \cmark & \cmark & \textsc{local}  & 52.24 & \best{79.63} & \best{71.15} & 30.57 & \best{73.62} & \second{49.04} \\
\midrule
SALAD & \cmark & \xmark & \textsc{salad}  & \second{57.37} & 73.33 & 56.09 & \second{43.95} & 66.98 & 38.22 \\
SALAD & \cmark & \xmark & \textsc{local}  & 38.14 & 73.31 & 52.88 & 19.75 & 67.43 & 36.31 \\
\midrule
SALAD & \cmark & \cmark & \textsc{salad}  & \best{61.54} & 77.04 & 66.67 & \best{46.50} & 70.01 & 43.31 \\
SALAD & \cmark & \cmark & \textsc{local}  & 48.40 & 78.77 & 67.95 & 28.66 & \second{73.50} & \best{49.68} \\
\rowcolor{chosengray}
SALAD & \cmark & \cmark & \textsc{hybrid}$^\star$ & \best{61.54} & 78.77 & 67.95 & \best{46.50} & \second{73.50} & \best{49.68} \\
\bottomrule
\end{tabular}
\end{table*}

We ablate the main design choices of CrossWeave in~\Cref{tab:ablation}. Specifically, we study the effect of the local patch-point alignment loss $\mathcal{L}_{p2p}$, the image-point-cloud aggregation loss $\mathcal{L}_{i2pc}$, the aggregation used during training (mean pooling or SALAD), and the choice of retrieval features. We compare three feature variants: \textsc{salad}, which uses the aggregated descriptor produced by the SALAD module; \textsc{local}, which mean-pools the patch and point features; and \textsc{hybrid}, which uses \textsc{salad} features for scene retrieval and \textsc{local} features for instance retrieval.

\paragraph{Local patch-point alignment improves both scene and instance retrieval}
Comparing the SALAD baseline trained only with $\mathcal{L}_{i2pc}$ against CrossWeave trained with both $\mathcal{L}_{i2pc}$ and $\mathcal{L}_{p2p}$ shows the benefit of local cross-modal supervision. On ScanNet, using \textsc{salad} features, CrossWeave improves scene Top-1 recall from $57.37$ to $61.54$, instance Top-1 recall from $73.33$ to $77.04$, and instance $R@75\%$ from $56.09$ to $66.67$. A similar trend is observed on 3RScan, where scene Top-1 recall improves from $43.95$ to $46.50$, instance Top-1 recall from $66.98$ to $70.01$, and instance $R@75\%$ from $38.22$ to $43.31$. These gains indicate that sparse local patch-point correspondences provide a stronger training signal than global image-point-cloud supervision alone, especially for fine-grained instance retrieval.

\paragraph{Local alignment alone is not sufficient for scene retrieval}
Training with only $\mathcal{L}_{p2p}$ and mean-pooled \textsc{local} features already gives strong instance retrieval, with $78.96$ and $73.30$ instance Top-1 recall on ScanNet and 3RScan. Scene retrieval, however, drops to $47.12$ and $27.39$ Top-1 recall, compared to $61.54$ and $46.50$ for the main model. Local alignment therefore learns discriminative object-level features, but does not by itself produce descriptors that separate scenes. Adding $\mathcal{L}_{i2pc}$ while keeping mean pooling improves scene Top-1 recall from $47.12$ to $52.24$ on ScanNet and from $27.39$ to $30.57$ on 3RScan, so the aggregated objective provides supervision that the local loss does not. Replacing mean pooling with SALAD further improves scene Top-1 recall to $61.54$ and $46.50$. Hence, both the aggregation objective and the aggregation module contribute to scene retrieval. Mean pooling with both losses gives the highest instance Top-1 recall ($79.63$ and $73.62$), slightly above the main model ($78.77$ and $73.50$), but its scene Top-1 recall is $9.30$ and $15.93$ points lower.

\paragraph{Aggregated features are better suited for scene retrieval}
For scene-level retrieval, \textsc{salad} features consistently outperform \textsc{local} features. Without $\mathcal{L}_{p2p}$, the SALAD baseline achieves $57.37$ Top-1 scene recall on ScanNet, compared to $38.14$ when using local features. On 3RScan, the gap is even larger, with \textsc{salad} achieving $43.95$ compared to $19.75$ for \textsc{local}. The same pattern holds for CrossWeave: \textsc{salad} features achieve $61.54$ scene Top-1 recall on ScanNet and $46.50$ on 3RScan, whereas \textsc{local} features achieve only $48.40$ and $28.66$, respectively. This confirms that scene retrieval benefits from a parametric aggregation module over mean-pooled local features.

\paragraph{Local features are better suited for instance retrieval} In contrast, instance retrieval benefits more from preserving local feature structure. For CrossWeave on ScanNet, using \textsc{local} features improves instance Top-1 recall from $77.04$ to $78.77$ and R@75\% from $66.67$ to $67.95$ compared to \textsc{salad} features. On 3RScan, the improvement is larger: instance Top-1 recall increases from $70.01$ to $73.50$, and $R@75\%$ improves from $43.31$ to $49.68$. This suggests that aggregation can discard fine-grained spatial and geometric cues that are useful for distinguishing object instances, whereas local descriptors retain the detailed correspondence information required for accurate instance matching.

\paragraph{The hybrid feature strategy gives the best overall trade-off} The \textsc{hybrid} configuration combines the strengths of both feature types by using \textsc{salad} descriptors for scene retrieval and \textsc{local} for instance retrieval. This yields the best trade-off between scene and instance retrieval across both datasets, and we therefore use this configuration as our main model.

\section{Conclusion}
\label{sec:conclusion}
In this work, we introduced CrossWeave, a novel framework for cross-modal scene and instance retrieval that demonstrates the superiority of sparse, local feature alignment over traditional global scene encoding. By pairing a local patch-to-point contrastive objective with optimal transport-based aggregation, CrossWeave achieves state-of-the-art scene and instance retrieval on ScanNet and 3RScan with a single model, trained without scene- or instance-level contrastive objectives. CrossWeave provides a flexible alternative for bridging the modality gap in spatial reasoning tasks.

\begin{ack}
Aadith Warrier is supported by the Reliance Foundation Postgraduate Scholarship.
\end{ack}

\clearpage
{
    \small
    \bibliographystyle{plainnat}
    \bibliography{references}
}

\newpage
\appendix
\crefalias{section}{appendix}
\crefalias{subsection}{appendix}
\raggedbottom

\begin{center}
    {\LARGE\bfseries Appendix}
\end{center}
\vspace{0.5em}

\section{Data Pre-processing and Training Optimization Details} \label{sec:more_implementation_details}
Both datasets are preprocessed by temporally subsampling the RGB-D frames: one in every 30 frames for ScanNet and one in every 5 frames for 3RScan, chosen according to the typical sequence length of each dataset. This yields 63,683 training images from ScanNet and 56,883 from 3RScan, for a total of 120,566 training images. For scene-level evaluation we use the 312 ScanNet and 157 3RScan validation scans. For instance-level evaluation we use 6,419 ScanNet and 3,274 3RScan validation instances.

To reduce dataloader overhead, projected pixel-point correspondences are precomputed and cached as \texttt{.pt} files, while FPS downsampling and patch-point pair generation are performed online. Utonia point features are also cached to avoid redundant forward passes over the same point clouds during training.

\section{Data Scaling}
\begin{figure}[H]
    \centering
    \includegraphics[width=0.5\linewidth]{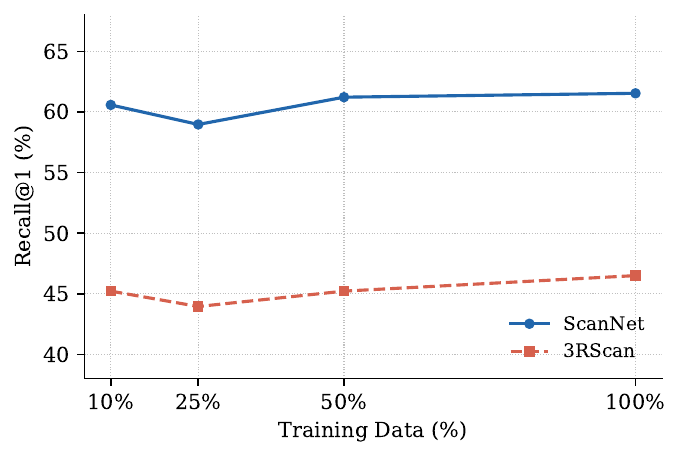}
    \caption{Scene retrieval performance (Top-1 recall) on ScanNet and 3RScan as a function of training data volume. Performance is stable across data regimes, with even 10\% of the full training set sufficient to outperform the previous state of the art.}
    \label{fig:data_scaling}
\end{figure}

We study scene retrieval performance as a function of training data volume. The results reported in the main paper use 120,566 images from ScanNet and 3RScan as described in \Cref{sec:more_implementation_details}. To study the effect of data scaling, we subsample the training set by uniform striding. As shown in~\Cref{fig:data_scaling}, CrossWeave is stable across data regimes: even at 10\% of the full training set, the model outperforms the previous state of the art. We attribute the data efficiency of our model to two factors: feature-space FPS during patch-point pair selection, which ensures semantic diversity within each batch, and the use of strong pre-trained encoders for each modality, which provide rich local features from the outset. Together, these properties mean that even a small number of uniformly sampled images contains sufficient supervision signal to learn a discriminative cross-modal representation.

\section{Effect of Scaling Camera Views}

\begin{figure}[H]
\centering
\begin{tikzpicture}[x=0.2cm, y=0.09cm, font=\footnotesize]
\foreach \v/\l in {5/30, 15/40, 25/50, 35/60, 45/70} {
    \draw[black!12] (1,\v) -- (32,\v);
    \node[left, text=black!70] at (1,\v) {\l};
}
\draw[black!50] (1,0) -- (32,0);
\foreach \x in {3,5,10,15,20,25,30} {
    \draw[black!50] (\x,0) -- (\x,-1);
    \node[below, text=black!70] at (\x,-1) {\x};
}
\draw[black!40, dashed] (10,0) -- (10,45);
\node[above, text=black!60, font=\scriptsize] at (10,45) {main results};
\draw[scannetblue, line width=1pt] plot[mark=*, mark size=1.6pt]
    coordinates {(3,20.83) (5,28.53) (10,36.54) (15,35.58) (20,37.18) (25,38.46) (30,39.10)};
\draw[rscanred, line width=1pt] plot[mark=square*, mark size=1.5pt]
    coordinates {(3,3.66) (5,13.22) (10,21.50) (15,24.04) (20,23.41) (25,25.96) (30,24.68)};
\node[right, text=black!80] at (30.5,39.10) {ScanNet};
\node[right, text=black!80] at (30.5,24.68) {3RScan};
\draw[scannetblue, line width=1pt] plot[mark=*, mark size=1.6pt] coordinates {(21,10) (23.5,10)};
\node[right, text=black!80] at (23.8,10) {ScanNet};
\draw[rscanred, line width=1pt] plot[mark=square*, mark size=1.5pt] coordinates {(21,5) (23.5,5)};
\node[right, text=black!80] at (23.8,5) {3RScan};
\node[below] at (16.5,-6) {Number of query views};
\node[rotate=90] at (-3.5,22.5) {Top-1 recall (\%)};
\end{tikzpicture}
\caption{Scene retrieval performance (Top-1 recall) as a function of the number of query camera views. The main results use 10 views.}
\label{fig:num_views}
\end{figure}
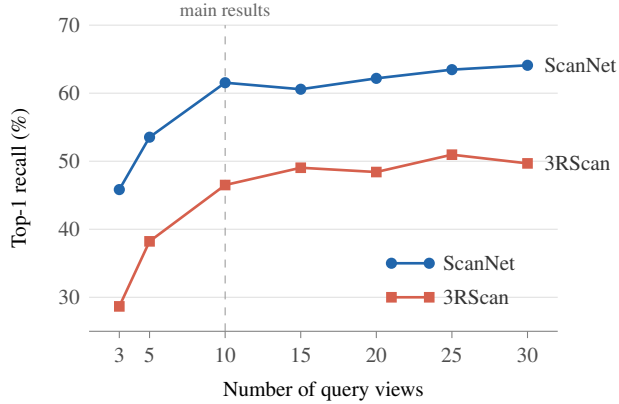

\Cref{fig:num_views} reports scene retrieval performance as a function of the number of query camera views at inference. Top-1 recall improves from 45.83 to 64.10 on ScanNet and from 28.66 to 49.68 on 3RScan as the number of views increases from 3 to 30. Most of the gain comes between 3 and 10 views; beyond 10 views, recall still improves by up to 2.56 points on ScanNet and 4.46 points on 3RScan. This behavior is a direct consequence of the architecture: because the SALAD aggregator maps a variable number of input features into a fixed-dimensional descriptor, CrossWeave naturally incorporates additional views without any retraining or architectural modification. In contrast, CrossOver~\cite{Sarkar_2025_CVPR} averages per-frame features with equal weight, and its authors report that performance stabilizes after 10 frames, with additional frames giving only slight improvements. CrossWeave continues to improve up to 25--30 views, since SALAD assigns patch features to clusters rather than averaging them. The ability to leverage more views at inference is particularly valuable in practical deployments, where a robot or AR device may capture varying numbers of observations depending on the environment.

\FloatBarrier

\section{Performance with Degraded Instance Masks}
\label{subsec:degraded_masks}

The instance retrieval results in \Cref{tab:scannet_3rscan_instance_results} use ground-truth instance masks for both the query images and the database point clouds. In deployment, these masks come from a segmentation model and contain errors. We therefore evaluate the main model, with \textsc{local} features for instance retrieval, under noisy 2D masks, noisy 3D masks, and both. For 2D, we generate masks with MobileSAM~\cite{mobile_sam}, prompted with a single point sampled well inside the ground-truth mask to mimic a human click, and keep the predicted mask with the highest IoU to the ground truth. For 3D, we corrupt each ground-truth instance point cloud in two ways: points from neighboring background and objects bleed in at the instance boundary, and contiguous chunks of the instance are dropped. We set the corruption so that the IoU degradation is similar to that of the 2D masks. Results for each mask condition are reported in \Cref{tab:degraded_masks}.

\begin{table}[H]
\centering
\caption{Instance retrieval under noisy instance masks, using the main model with \textsc{local} features.}
\label{tab:degraded_masks}
\footnotesize
\setlength{\tabcolsep}{6pt}
\renewcommand{\arraystretch}{1.15}
\begin{tabular}{ll cc ccc}
\toprule
& & \multicolumn{2}{c}{Instance Retrieval} & \multicolumn{3}{c}{Scene-Level Recall} \\
\cmidrule(lr){3-4} \cmidrule(lr){5-7}
\textbf{Dataset} & \textbf{Mask condition} & Top-1 & Top-3 & R@25\% & R@50\% & R@75\% \\
\midrule
\multirow{4}{*}{ScanNet}
& GT            & 78.77 & 94.05 & 98.40 & 88.78 & 67.95 \\
& 2D noisy      & 75.75 & 93.02 & 97.44 & 88.14 & 62.50 \\
& 3D noisy      & 75.33 & 92.91 & 97.12 & 87.50 & 63.46 \\
& 2D + 3D noisy & 72.29 & 91.86 & 97.12 & 85.90 & 55.45 \\
\midrule
\multirow{4}{*}{3RScan}
& GT            & 73.50 & 92.73 & 98.09 & 92.99 & 49.68 \\
& 2D noisy      & 68.00 & 89.95 & 97.45 & 88.54 & 34.39 \\
& 3D noisy      & 65.65 & 88.79 & 96.82 & 84.08 & 32.48 \\
& 2D + 3D noisy & 61.28 & 86.11 & 95.54 & 76.43 & 26.75 \\
\bottomrule
\end{tabular}
\end{table}

\FloatBarrier

\section{FPS Sampling Visualizations}

\begin{figure}[H]
    \centering
    \begin{subfigure}[t]{0.32\linewidth}
        \centering
        \includegraphics[width=\linewidth]{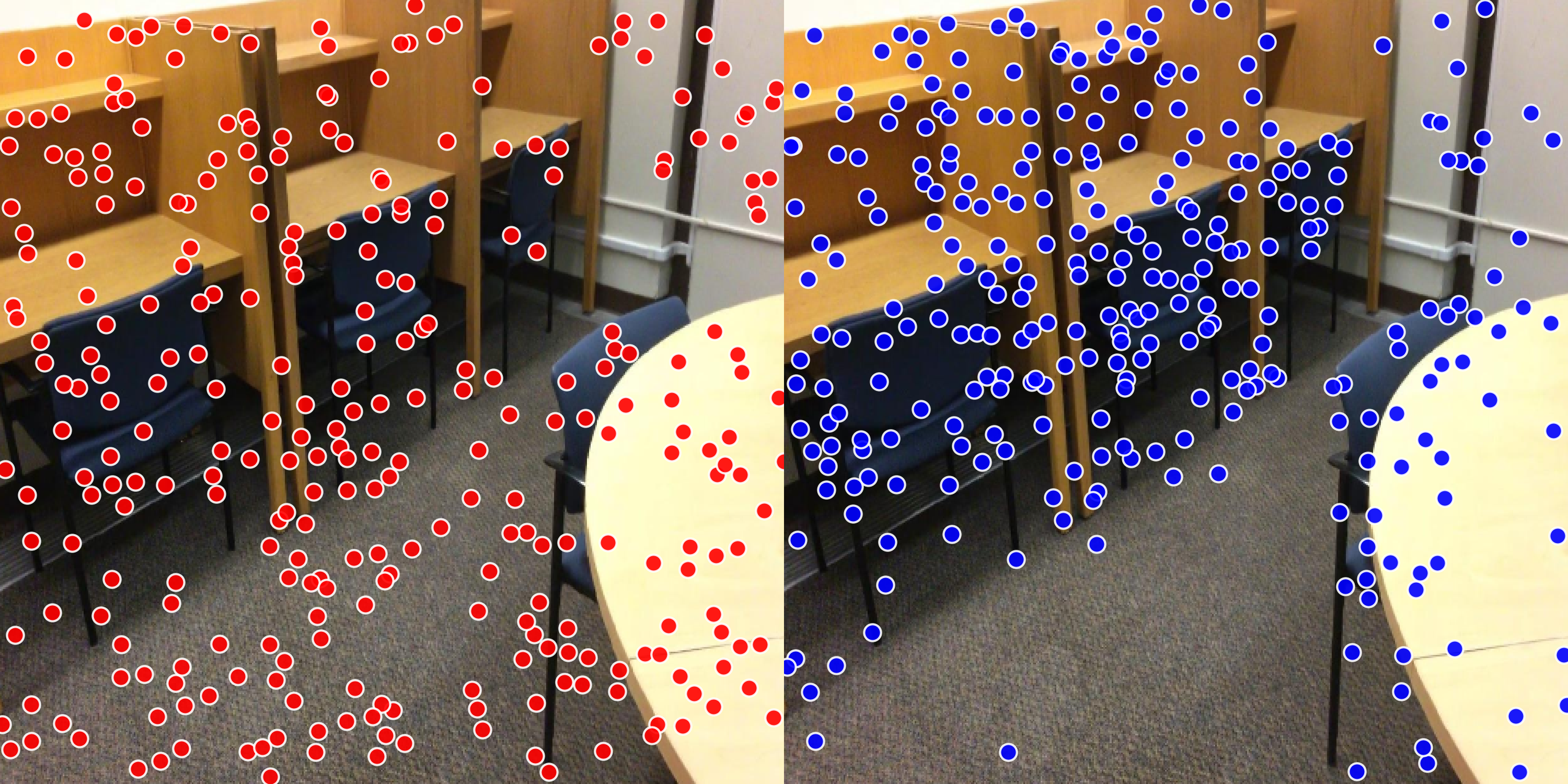}
        \caption{ScanNet \texttt{scene0665\_00}}
        \label{fig:sub1}
    \end{subfigure}
    \hfill
    \begin{subfigure}[t]{0.32\linewidth}
        \centering
        \includegraphics[width=\linewidth]{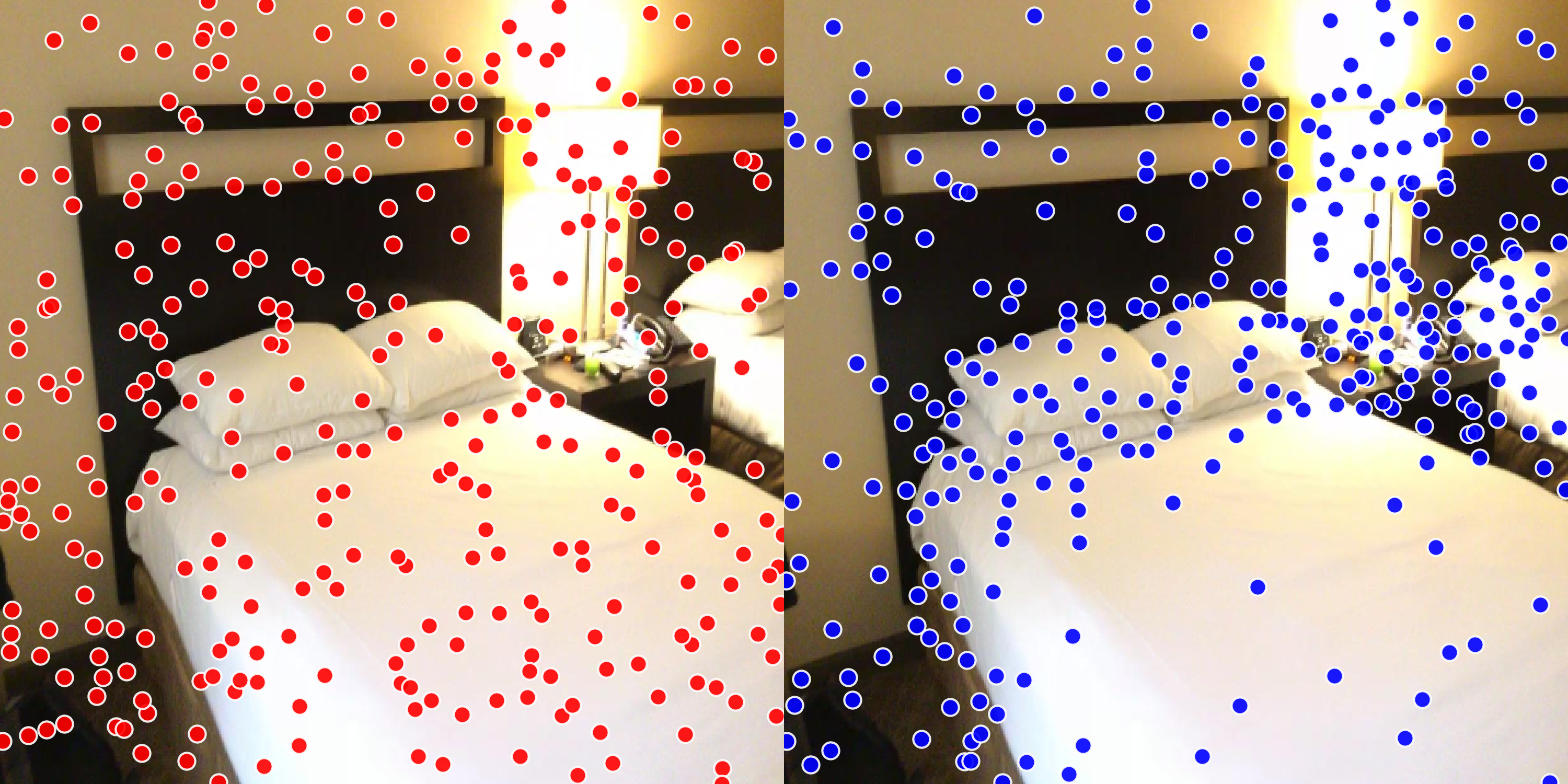}
        \caption{ScanNet \texttt{scene0435\_02}}
        \label{fig:sub2}
    \end{subfigure}
    \hfill
    \begin{subfigure}[t]{0.32\linewidth}
        \centering
        \includegraphics[width=\linewidth]{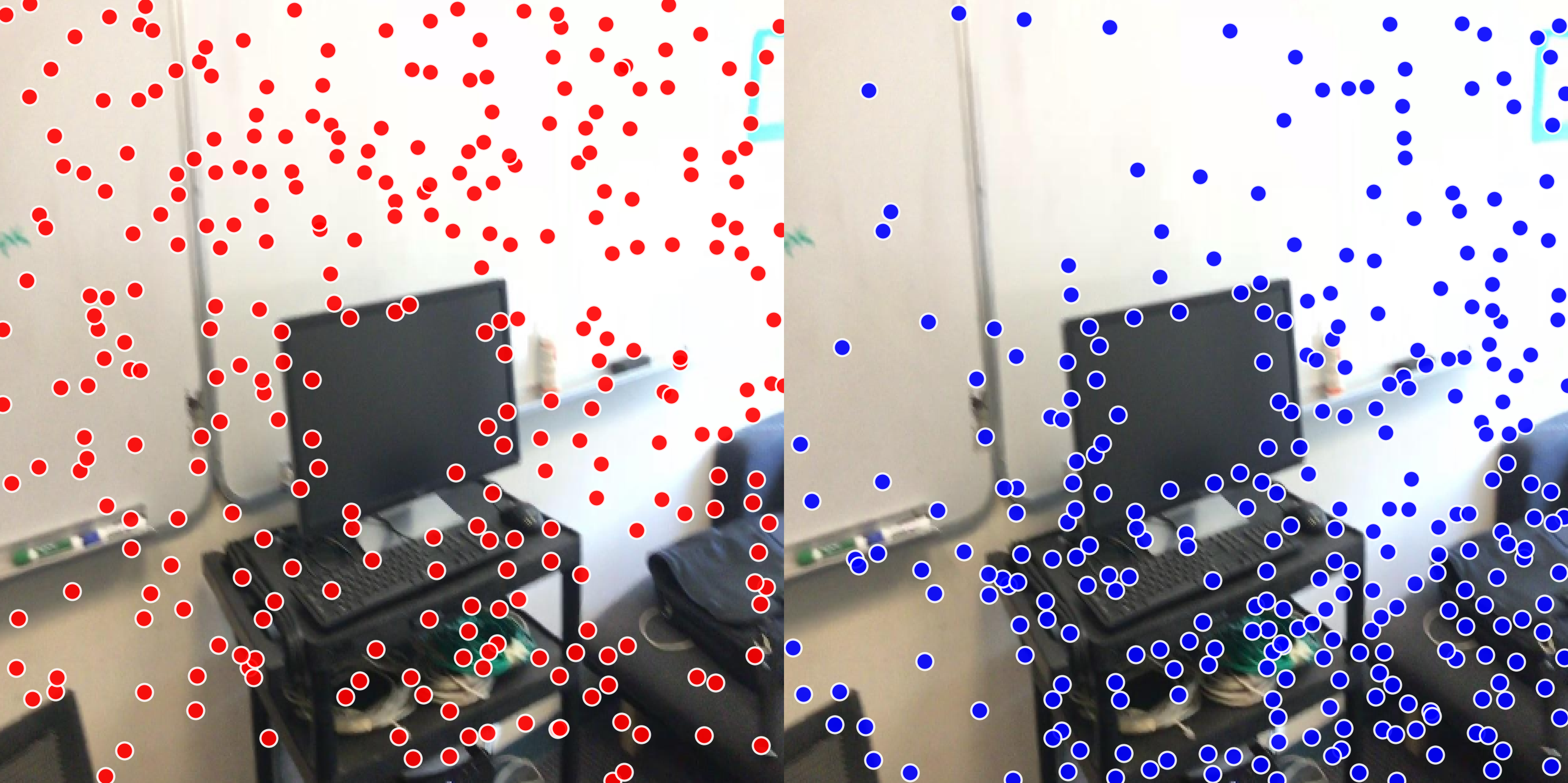}
        \caption{ScanNet \texttt{scene0474\_00}}
        \label{fig:sub3}
    \end{subfigure}
    \caption{Visualizations for the effect of FPS sampling. Images with red dots denote random sampling. Images with blue dots denote FPS sampled points. It can clearly be observed that FPS enables us to avoid over-sampling on the often visually dominant but geometrically uninformative areas in the images such as walls, floors and other flat surfaces, thereby boosting feature diversity within each training batch. Note: We have visualized a large number of pixel-point pairs in these images; in our training framework we only use 32 patch-point pairs. Refer to \Cref{subsec:training} for more details.}
    \label{fig:main}
\end{figure}

\FloatBarrier

\section{Additional Visualizations for Qualitative Results} \label{sec:addl_viz}

\subsection{ScanNet Visualizations}
\begin{figure}[H]
    \centering
    \includegraphics[width=\linewidth]{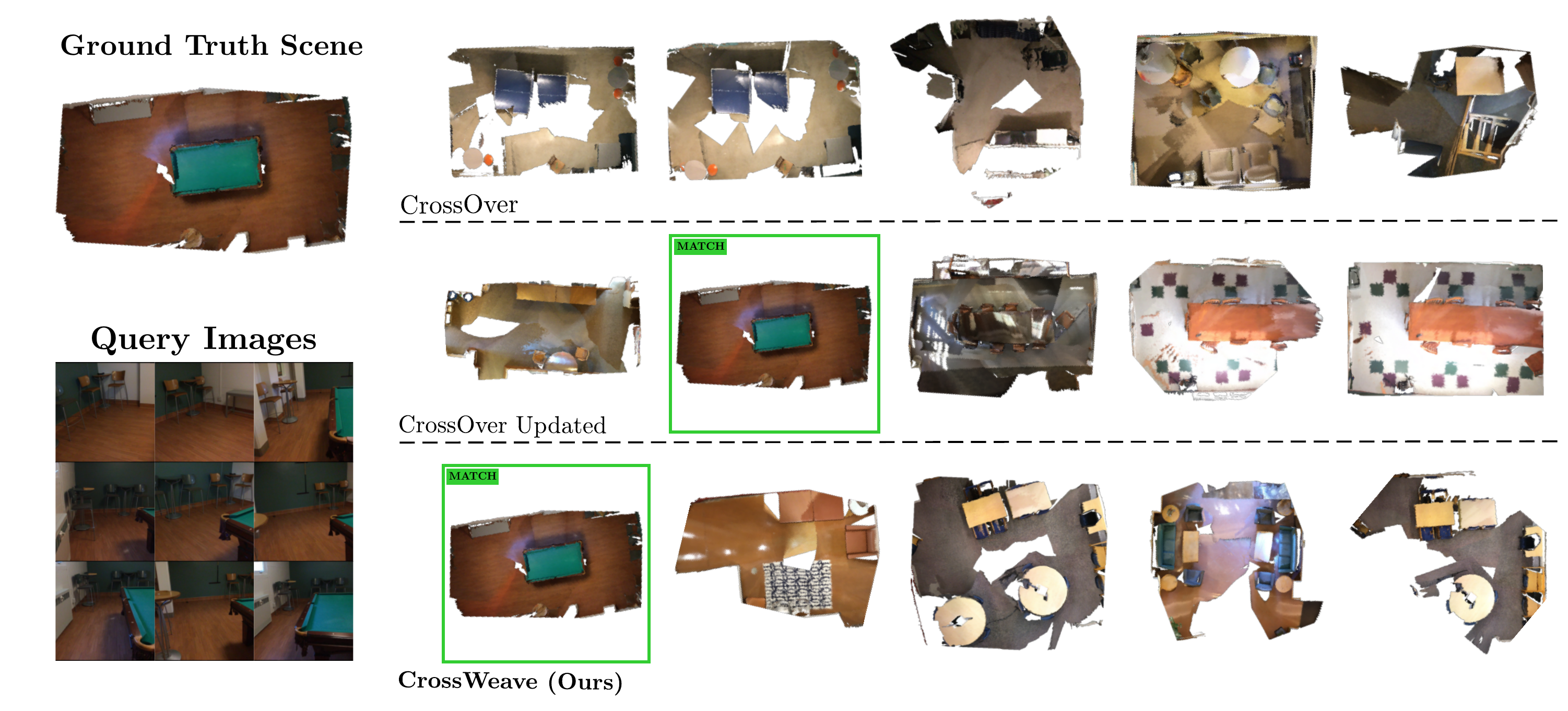}
    \caption{Comparisons of Top-5 retrievals on a ScanNet room scene. Retrievals are ordered from most to least similar moving left to right, with correct matches highlighted in green. CrossWeave successfully retrieves the correct scene at rank 1, while CrossOver (Updated) retrieves it at rank 2, and the base model of CrossOver fails within the top 5.}
    \label{fig:scannet_qual_result1_supplementary}
\end{figure}

\begin{figure}[H]
    \centering
    \includegraphics[width=\linewidth]{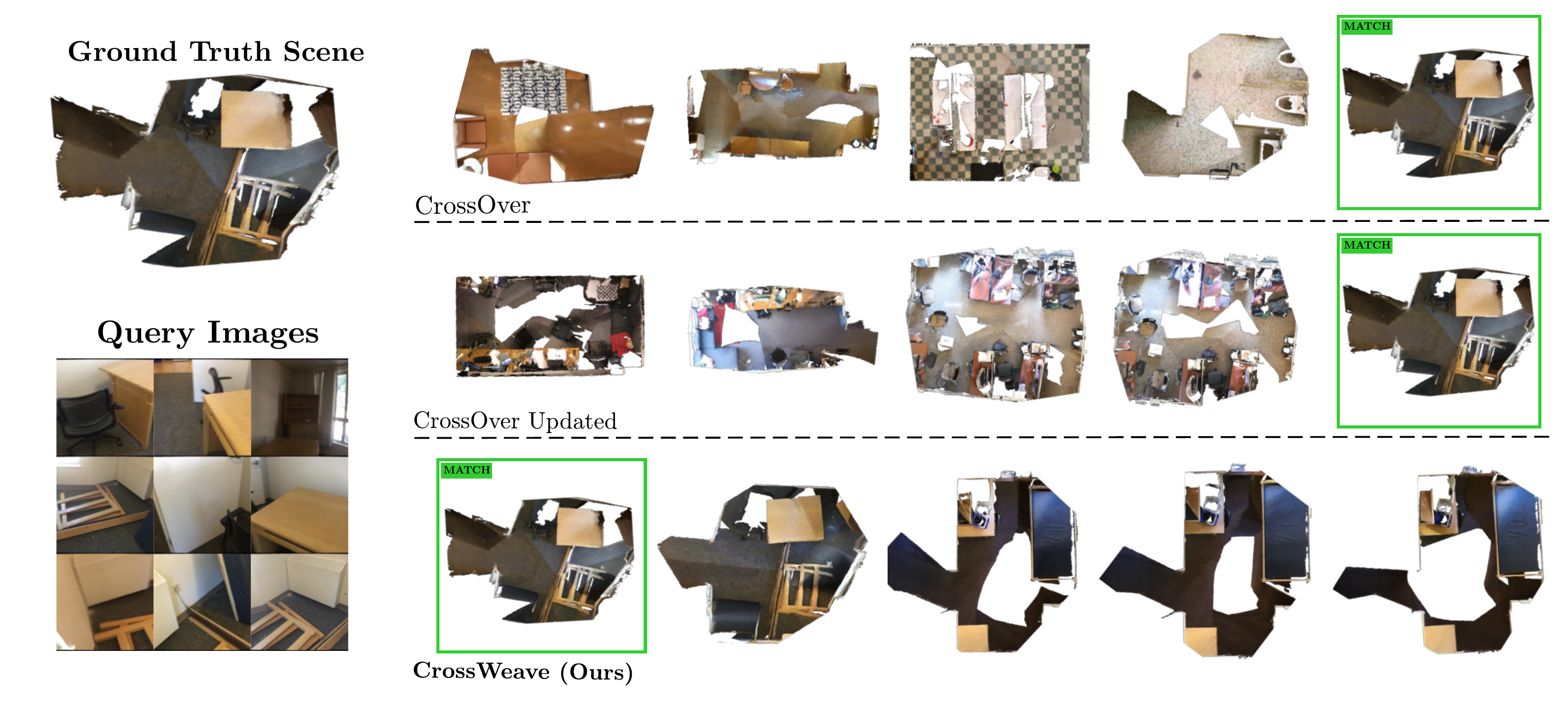}
    \caption{Additional qualitative results on a ScanNet office scene. CrossWeave retrieves the ground truth point cloud as its top match, whereas both CrossOver (Updated) and CrossOver only identify the correct scene at rank 5.}
    \label{fig:scannet_qual_result2_supplementary}
\end{figure}

\FloatBarrier

\subsection{3RScan Visualizations}
\begin{figure}[H]
    \centering
    \includegraphics[width=\linewidth]{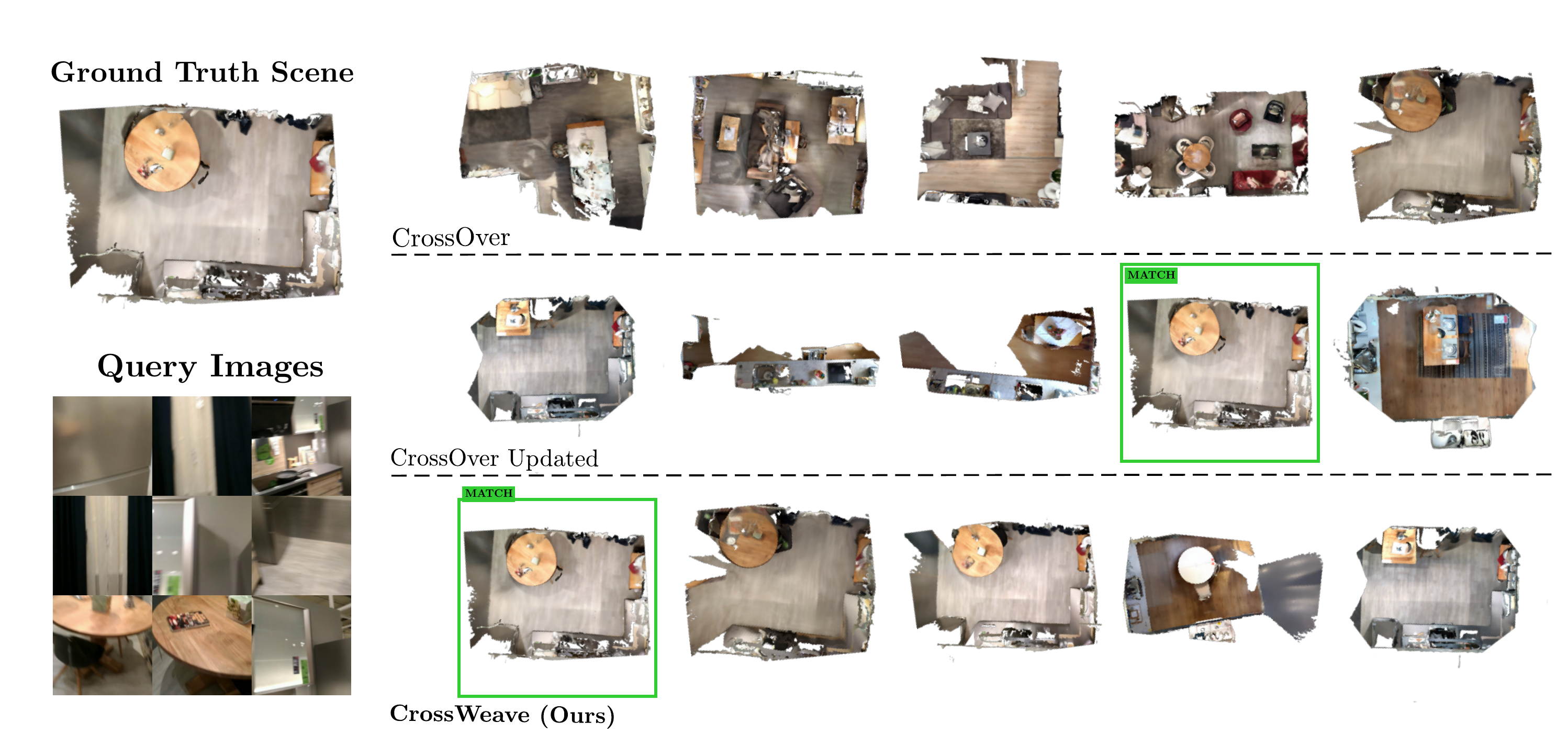}
    \caption{Qualitative results on a 3RScan dining area scene. CrossWeave successfully retrieves the ground truth point cloud as its top match, while CrossOver (Updated) ranks it fourth and the base CrossOver model fails to retrieve it within the top 5.}
    \label{fig:3rscan_qual_result1_supplementary}
\end{figure}

\begin{figure}[H]
    \centering
    \includegraphics[width=\linewidth]{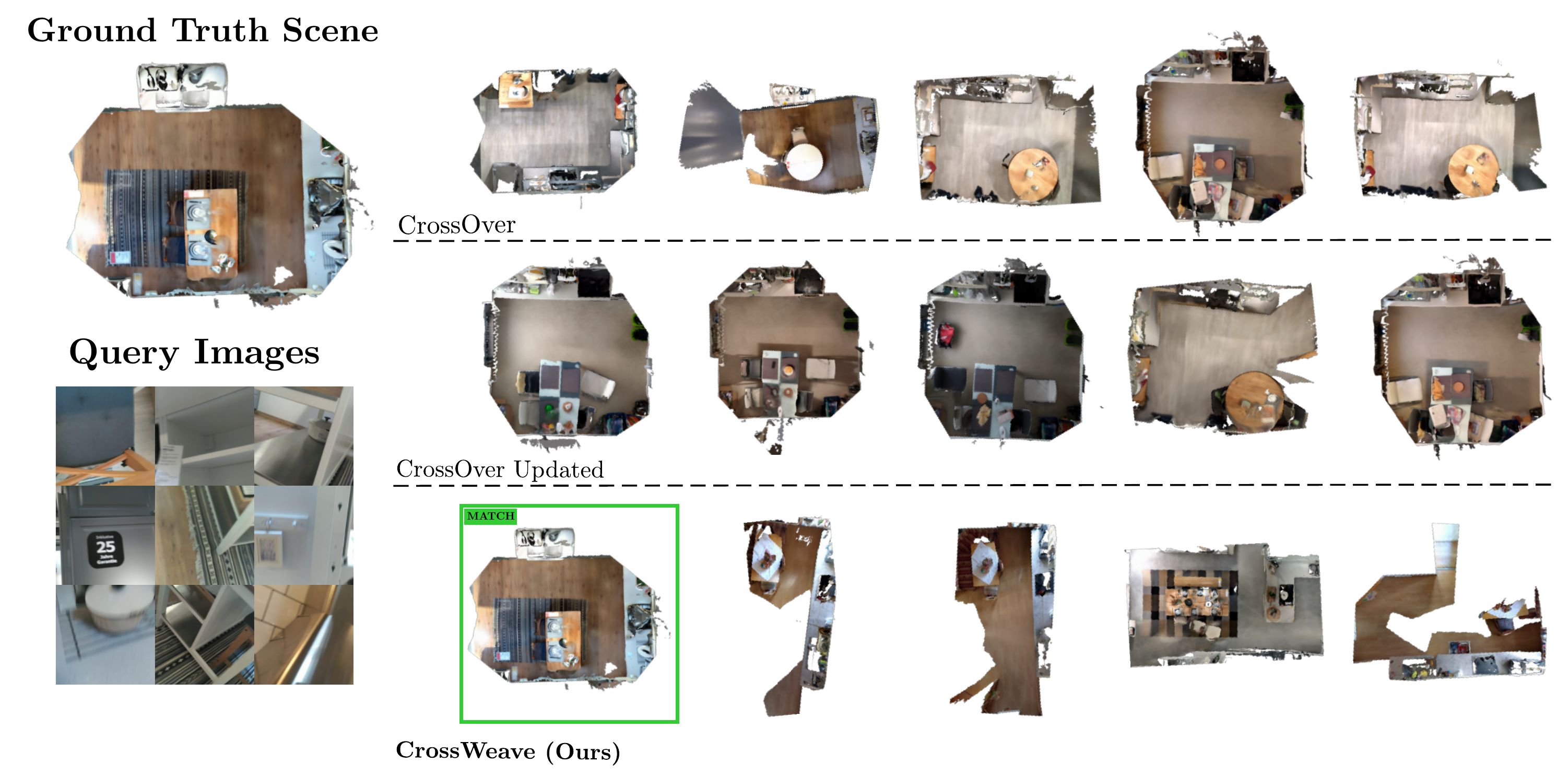}
    \caption{Retrieval results on a 3RScan kitchen and dining scene. CrossWeave demonstrates superior discriminative ability by retrieving the exact scene at rank 1 from highly localized query images, whereas both baseline models completely fail to place the correct match within their top 5 retrievals.}
    \label{fig:3rscan_qual_result2_supplementary}

    \vspace{2em}

    \includegraphics[width=\linewidth]{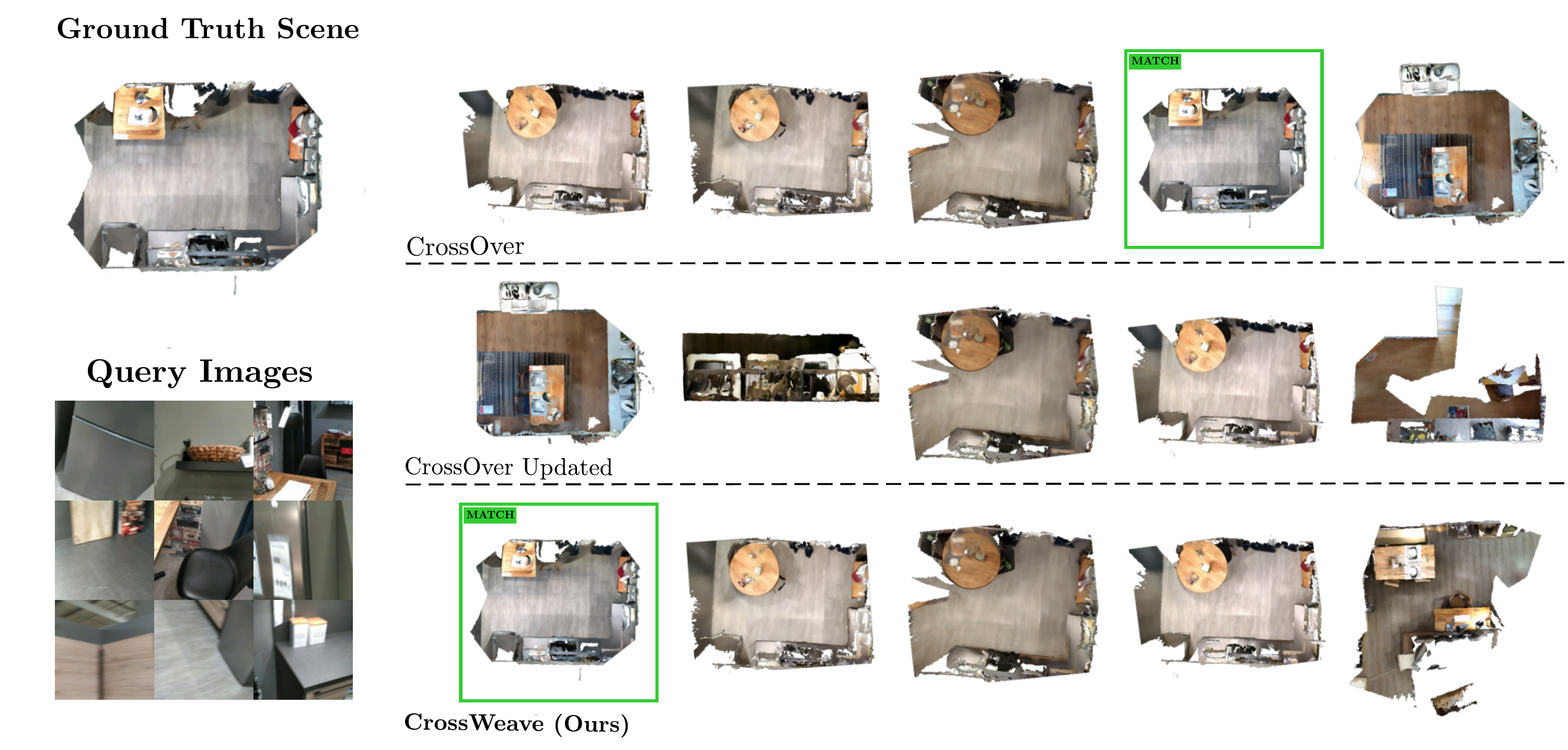}
    \caption{Additional results on a 3RScan kitchen scene. CrossWeave correctly identifies the target environment at rank 1, outperforming CrossOver, which places the match at rank 4, and CrossOver (Updated), which misses it entirely.}
    \label{fig:3rscan_qual_result3_supplementary}
\end{figure}

\FloatBarrier
\clearpage

\begin{figure}[H]
    \centering
    \includegraphics[width=\linewidth]{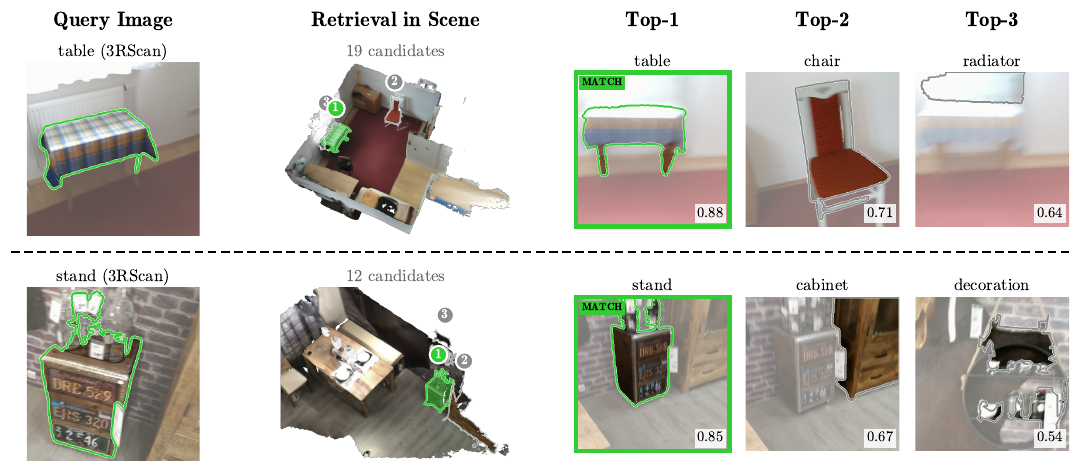}
    \caption{Qualitative instance retrieval on 3RScan. For each query view with its instance mask (left), CrossWeave ranks the instances of the same scene (middle) by cosine similarity. The Top-3 retrievals are shown with their similarity scores, and the correct match is outlined in green.}
    \label{fig:3rscan_instance_retrieval_supplementary}
\end{figure}

\section{Failure Case Analysis} \label{sec:failure}
\begin{figure}[H]
    \centering
    \includegraphics[width=\linewidth]{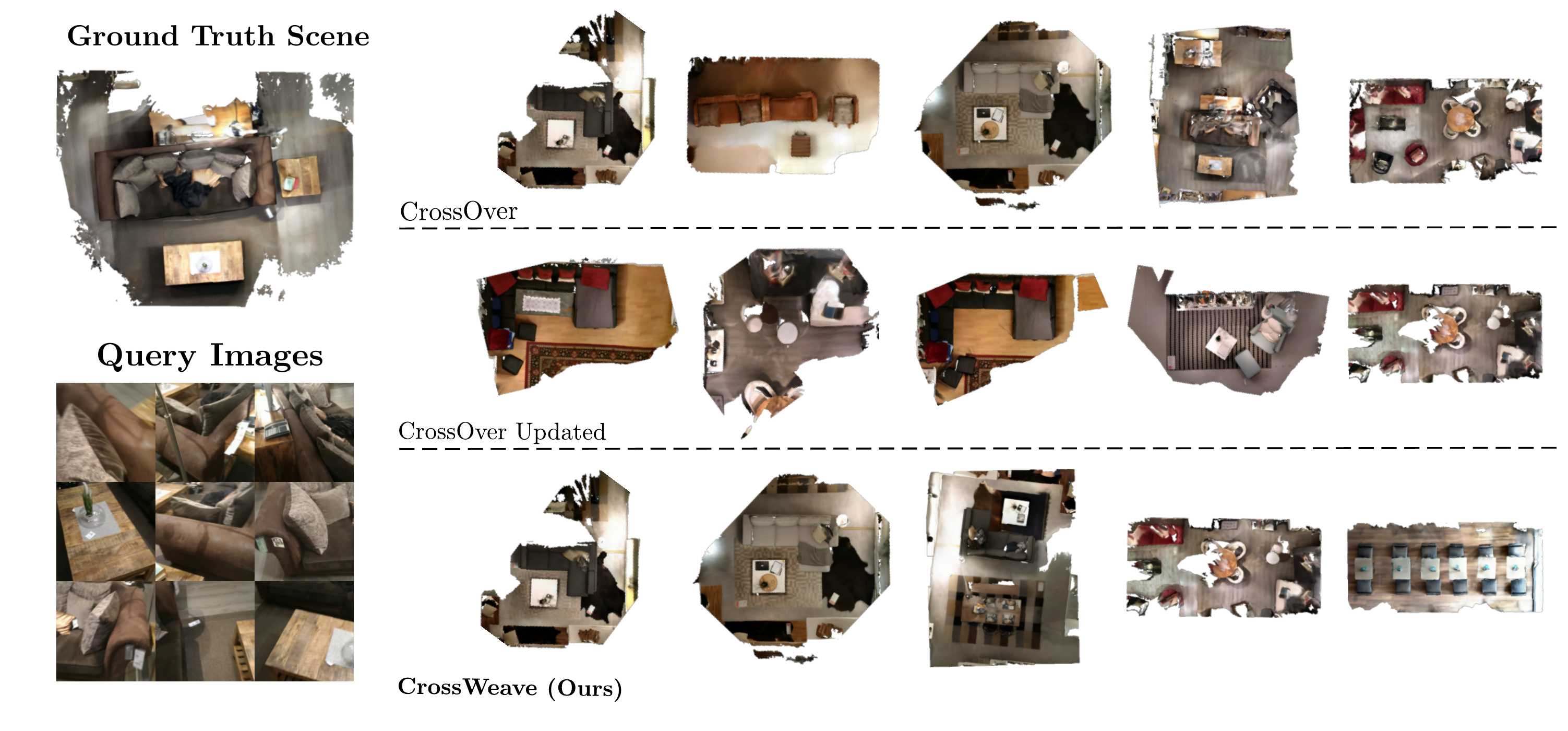}
    \caption{Failure case on a 3RScan living room scene. All three models miss the ground-truth scene in the top 5, as the retrieved scenes are visually and structurally similar to the query (a brown sectional sofa with a wooden coffee table). CrossWeave's top matches still share the furniture arrangement (sofas and central coffee tables), whereas CrossOver (Updated) retrieves structurally different scenes.}
    \label{fig:failure_case_analysis}
\end{figure}

\begin{figure}[H]
    \centering
    \includegraphics[width=\linewidth]{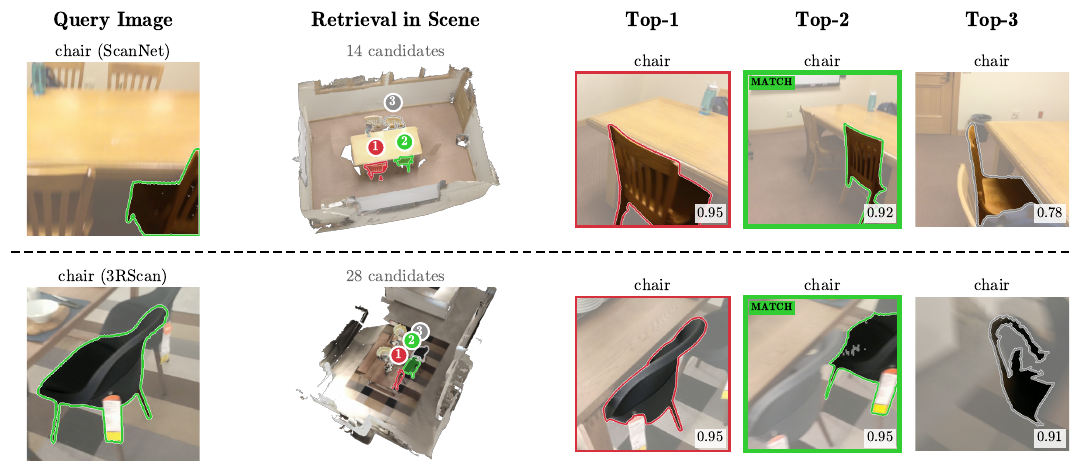}
    \caption{Instance retrieval failures on ScanNet and 3RScan. In both cases the query is a chair and the Top-1 retrieval (red) is another chair of the same type in the same scene, while the correct instance (green) is ranked second with a similar score.}
    \label{fig:instance_failure_case_analysis}
\end{figure}

\FloatBarrier

\section{Limitations} \label{sec:limitations}
Despite strong performance on most benchmarks, CrossWeave has several limitations. First, at inference, CrossWeave assumes that query images are captured within a single viewing session; queries that span multiple sessions are not explicitly handled. Second, the training correspondences are derived from the camera intrinsics, camera poses, depth rendered from the reconstructed mesh, and the reconstruction itself. Errors in calibration or pose, missing or inaccurate rendered depth, and incomplete or noisy reconstructions produce incorrect or missing patch-point pairs. We use the reconstructions and poses provided with ScanNet and 3RScan and do not study the effect of such errors. Finally, instance retrieval requires segmented instances, both as query masks and as database point clouds, making retrieval quality dependent on the upstream segmentation, which is outside the scope of this work. \Cref{subsec:degraded_masks} shows that retrieval degrades under noisy 2D and 3D masks, most strongly at R@75\%.

\section{Broader Impacts} \label{sec:broader_impacts}
CrossWeave enhances spatial intelligence for indoor robotics, autonomous navigation, and mixed-reality systems through efficient cross-modal localization. However, by relying on pre-trained foundation models (DINOv2 and Utonia), it risks inheriting and propagating their latent training biases. This could result in disparate retrieval performance across culturally or socioeconomically diverse architectural styles, causing the system to underperform in underrepresented environments. Real-world deployments must actively audit these underlying models to mitigate biased spatial reasoning.

\end{document}